\documentclass[letterpaper, 10 pt, conference]{ieeeconf}  

\IEEEoverridecommandlockouts                              

\usepackage{graphics} 
\usepackage{epsfig} 
\usepackage{mathptmx} 
\usepackage{times} 
\usepackage{amsmath} 
\usepackage{amssymb}  
\usepackage{tikz}

\usepackage{cite}

\makeatletter
\def\@citex[#1]#2{\leavevmode
\let\@citea\@empty
\@cite{\@for\@citeb:=#2\do
{\@citea\def\@citea{,\penalty\@m\ }%
\edef\@citeb{\expandafter\@firstofone\@citeb\@empty}%
\if@filesw\immediate\write\@auxout{\string\citation{\@citeb}}\fi
\@ifundefined{b@\@citeb}{\hbox{\reset@font\bfseries ?}%
\G@refundefinedtrue
\@latex@warning
{Citation `\@citeb' on page \thepage \space undefined}}%
{\@cite@ofmt{\csname b@\@citeb\endcsname}}}}{#1}}
\makeatother

\newcommand\Tstrut{\rule{0pt}{2.0ex}}         
\newcommand\Bstrut{\rule[-0.9ex]{0pt}{0pt}} 
\newcommand\Mstrut{\rule[-0.0ex]{0pt}{0pt}}

\newcommand\copyrighttext{%
  \footnotesize This work has been submitted to the IEEE for possible publication. Copyright may be transferred without notice, after which this version may no longer be accessible.}%
\newcommand\copyrightnotice{%
\begin{tikzpicture}[remember picture,overlay]%
\node[anchor=south,yshift=10pt] at (current page.south) {\fbox{\parbox{\dimexpr\textwidth-2cm}{\copyrighttext}}};%
\end{tikzpicture}%
\vspace{-10pt}%
}

\title{\LARGE \bf
Semantic Segmentation of Video Sequences with Convolutional LSTMs
}

\author{Andreas Pfeuffer$^{1}$, Karina Schulz$^{1}$, and Klaus Dietmayer$^{1}$
\thanks{$^{1}$Andreas Pfeuffer, Karina Schulz, and Klaus Dietmayer are with the Institute of Measurement, Control, and Microtechnology, Ulm University, 89081 Ulm, Germany, firstname.lastname@uni-ulm.de}%
}

\begin{document}

\maketitle
\copyrightnotice
\thispagestyle{empty}
\pagestyle{empty}

\begin{abstract}

	Most of the semantic segmentation approaches have been developed for single image segmentation, and hence, video sequences are currently segmented by processing each frame of the video sequence separately. The disadvantage of this is that temporal image information is not considered, which improves the performance of the segmentation approach. One possibility to include temporal information is to use recurrent neural networks. 
	However, there are only a few approaches using recurrent networks for video segmentation so far. These approaches extend the encoder-decoder network architecture of well-known segmentation approaches and place convolutional LSTM layers between encoder and decoder. However, in this paper it is shown that this position is not optimal, and that other positions in the network exhibit better performance. Nowadays, state-of-the-art segmentation approaches rarely use the classical encoder-decoder structure, but use multi-branch architectures. These architectures are more complex, and hence, it is more difficult to place the recurrent units at a proper position. In this work, the multi-branch architectures are extended by convolutional LSTM layers at different positions and evaluated on two different datasets in order to find the best one. It turned out that the proposed approach outperforms the pure CNN-based approach for up to 1.6 percent. 

\end{abstract}


\section{INTRODUCTION}

\begin{figure}[t]
		\centering
		\includegraphics[width=1.0\columnwidth]{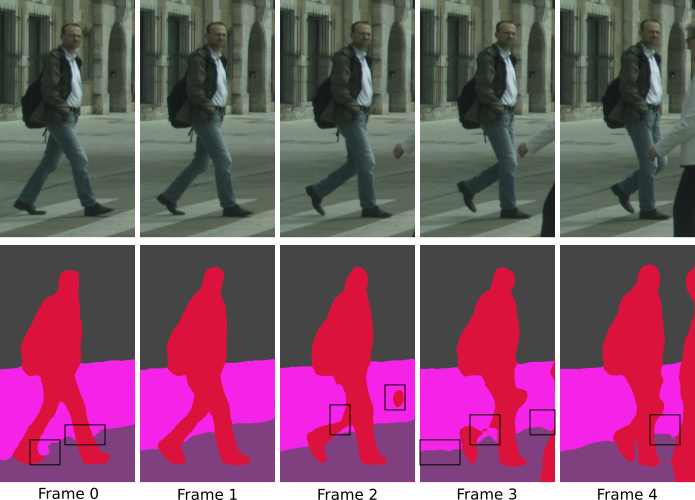}
		\caption{Segmentation map of a video sequence yielded by the ICNet. The black boxes show typical errors such as partly segmented objects, flickering edges, and flickering (ghost) objects}
		\label{fig_PedestrianSequence}
	\end{figure}

	A challenge of autonomous driving is to understand the environment as good as possible. Hence, multiple sensors are used in self-driving cars, such as the classical RGB camera. In order to reduce the flood of information of the camera, the images are segmented. Image segmentation denotes the task to assign each image pixel a predefined class, e.g. car, pedestrian, or road.  
	State-of-the-art approaches, such as PSPNet \cite{Zhao_2017_PyramidScenParsingNetwork} or DeepLab \cite{Chen_2017_DeepLab_SemanticImageSegmentationWithDeepConvolutionalNets_AtrousConvolution_andFullyConnectedCRFs}, are based on convolutional neural networks (CNNs) and achieve 
	very good results on several datasets.
	However, these approaches are not applicable in the case of autonomous driving, since the inference time for one image amounts to about one second and more, for instance, the PSPNet \cite{Zhao_2017_PyramidScenParsingNetwork} takes about $1.2$ seconds and the DeepLab v3+ \cite{Chen_2018_Deeplabv3p_EncoderDecoderWithAtrousSeparableConvolutionForSemanticImageSegmentation} about $5$ seconds on a Nvidia Titan X. In contrast, the performance of current real-time capable approaches, such as SegNet \cite{2015_Badrinarayanan_SegNet_ADeepConvolutionalEncoderDecoderARchitectureForImageSegmenatation}, ENet \cite{Paszke_2016_ENet_ADeepNeuralNetworkArchitectureForRealTimeSemanticSegmentation}, and ICNet \cite{Zhao_2017_ICNet_forRealTimeSemanticSegmentationOnHighResolutionImages}, is  much worse, and more errors occur. 
	Typical segmentation errors are blurred and flickering object edges, partly segmented objects, and flickering (ghost) objects.
	Many of these errors often only occur in a single frame of a video sequence, and are classified correctly in the next frame, as  shown in Fig \ref{fig_PedestrianSequence}, where a short video sequence was segmented by the ICNet. For instance, the pedestrian was classified correctly in the first to third and in the last frame of the video, while parts of the leg were not detected in the fourth frame. Furthermore, the border between road and sidewalk is flickering during the video, and a ghost object occurs at time step two.
	The described errors can be avoided by additionally considering image information of the previous frames instead of processing each image independently. One possibility to take account of the last frames is to use recurrent neural networks. They are able to store information of the past time steps and to reuse them in the current time step. Frequently used recurrent networks are Long-Short-Term Memory networks (LSTM) \cite{Hochreiter_1997_LongShortTermMemory}, which can be easily trained and integrated in comparison to other recurrent neural networks. An extension of LSTMs are convolutional LSTMs (convLSTMs \cite{Shi_2015_ConvolutionalLSTMNetwork_AMachineLearningApproachForPrecipitationNowcasting}), which are more suitable for image processing tasks. 
	
	Convolutional LSTM layers can be added at different positions in the network. For instance, they can be integrated directly in front of the softmax layer, which corresponds to a temporal filtering of the result. 
	Another possible location is between the encoder and decoder in the case of a encoder-decoder network architecture, and is motivated by the fact that the encoder extracts global features of the image. These global features should not change very much between two neighboring frames so that the usage of the previous global image features may improve the segmentation.	
	In this work, several positions of convLSTM layers in different, real-time capable, state-of-the-art semantic segmentation approaches are compared in terms of accuracy, mIoU and inference time on the Cityscapes dataset. It is also investigated if the LSTM based semantic segmentation approaches outperform the pure CNN based approaches, and which LSTM position in the network architecture delivers the best results in the case of semantic video segmentation.


\section{RELATED WORK}
\label{section_related_work}

	Nowadays, there are many approaches which deal with semantic segmentation. Todays approaches rely on neural networks and exhibit high accuracy. An overview of the best approaches with the highest accuracies can be found on the web pages of the well-known benchmarks such as the Cityscapes dataset \cite{cityscape_dataset}. However, only a few of these approaches are real-time capable, and hence, most of them are not applicable in safety-critical applications, such as self-driving cars. Moreover, most approaches are originally only developed for image segmentation and not for video segmentation, that means that each frame of a video sequence is processed independently, and temporal information of the video is not considered. 
	Though, the usage of temporal information might improve the accuracy, since errors occurring only at one frame, such as flickering (ghost) objects and flickering edges, might be avoided by means of these additional information. Hence, temporal information are used in this paper to improve the segmentation results. 
	Generally, there are different possibilities to include temporal information into neural networks. A naive way to do this is to concatenate the current frame with the previous $N$ frames to a common input tensor and to process these $N+1$ frames together with one common CNN. Previous works, such as \cite{Karpathy_2014_LargeScaleVideoClassificationWithConvolutionalNeuralNetworks}, show that the prediction improves only slightly by this procedure. 
	A more successful approach is to use recurrent neural networks (RNNs) and combine them with classical image segmentation approaches. In this section, an overview over these approaches and RNNs in general is given.
	
	Recurrent neural networks are neural networks containing loop connections. By means of these loops, RNNs can learn complex dynamics, so that sequential data can be processed. A great problem of RNNs is that their training is very difficult due to the vanishing and exploding gradients problem, and hence, they have not been widely used. 
	In 1997, Hochreiter and Schmidhuber introduced the Long-Short Term Memory networks (LSTMs \cite{Hochreiter_1997_LongShortTermMemory}), which overcomes the problem of vanishing/exploding gradients of the classical RNNs. The key innovation of the LSTMs compared to the RNNs is its memory cell, which is able to store state information. The LSTM cells can learn when to access, clear, and write to the memory during training and is able to memorize information over a long period.
	The LSTMs are now successfully applied in several applications, such as speech-recognition \cite{Liu_XXXX_DeepConvolutionalAndLSTMNeuralNetworksForAcousticModellingInAutomaticSpeechRecognition, Geiger_XXXX_RobustSpeechRecognitionUsingLongShortTermMemoryRecurrentNeuralNetworksForHybridAcousticModelling}, hand-writing recognition \cite{Graves_2013_GeneratingSequencesWithRecurrentNeuralNetworks} and machine translation \cite{Sutsekever_2014_SequenceToSequenceLearningWithNeuralNetworks}. However, they are not so suitable for image processing applications analogously to Multi Layer Perceptrons (MLPs), since they are translational variant and memory intensive due to large weight matrix sizes. Shi et al. refined the LSTMs and introduced the convLSTMs \cite{Shi_2015_ConvolutionalLSTMNetwork_AMachineLearningApproachForPrecipitationNowcasting} in 2015, which correspond to the CNNs in feedforward neural networks. Due to these advantages, ConvLSTMs become increasingly popular, and are used in more and more image processing applications. 
	
	In the case of semantic segmentation, there are two fields of application for RNNs. On one side, they are used to capture temporal information of a video sequence. On the other side, RNNs are used to learn the global context of one image. The image is divided into small regions or superpixels. Each of these image segments are fed one after another into the LSTM so that the network can learn the spatial relationship of the image regions. Examples are the so called multi-dimensional LSTM networks \cite{Byeon_2015_SceneLabelingWithLSTMRecurrentNeuralNetworks, Li_2016_RGBDSceneLabelingWithLongShortTermMemorizedFusionModel, Liang_2016_SemanticObjectParsingWithGraphLSTM}, and the work of Ren et al. \cite{Ren_2016_End2EndInstanceSegmentationAndCountingWithRecurrentAttention}, whose approach delivers a single instance at each iteration of the RNN.	
	The second field of application are networks using temporal information of video sequences. Until now, only very few approaches combining semantic segmentation approaches with recurrent structures have been proposed. One of the first approaches was the Recurrent Fully Convolutional Network (RFCN, \cite{Valipour_2017_RecurrentFullyConvolutionalNetworksForVideoSegmentation}), which was introduced by Valipour et al. in 2017. They extended the FCN approach \cite{Long_2015_FullyConvolutionalNetworksForSemanticSegmentation} by placing a recurrent unit between the encoder and the decoder, and exhibit better performance on the SegTrack, Davis, and Moving MNIST dataset. 
	In \cite{Yurdakul_2017_SemanticSegmentationOfRGBDVideosWithRecurrentFullyConvolutionalNeuralNetworks}, Yurdakul et al. evaluated different recurrent structures, such as convRNN, convGRU, and convLSTM, on the virtual Kitti dataset \cite{virtualKittiDataset}, where the encoder and decoder consist of modified VGG19 architectures \cite{Simonyan_2015_VeryDeepConvolutionalNetworksForLargeScaleImageRecognition}. They found out that the convLSTM performs best. However, the difference between the convLSTM-based approach and the pure CNN-based approach is not large - in some cases the LSTM-based approaches even performs worse. 
	In \cite{Nabavi_2018_FutureSemanticSegmentationWithConvolutionalLSTM}, the encoder-LSTM-decoder architecture was extended by skip-connections between encoder and decoder. Each of the skip-connections contains a convLSTM cell, too, and the encoder and decoder are based on the ResNet architecture \cite{Xie_2016_AggregatedResidualTransformationsForDeepNeuralNetworks}. However, the network was designed to predict the segmentation map of the future frame given the last $N$ segmentation maps.

	A commonality of present recurrent segmentation approaches is their encoder-decoder architecture, where convLSTM layers are placed between the encoder and the decoder part. However,  this position is not optimal, as results of this work show. Better results can be yielded if the convLSTM layer is located directly before the softmax layer. Furthermore, state-of-the-art approaches (e.g. \cite{Zhao_2017_ICNet_forRealTimeSemanticSegmentationOnHighResolutionImages, Zhao_2017_PyramidScenParsingNetwork}) do not often use the classical encoder-decoder structure any more, but use several branches, for example for different resolutions, and combine these branches at the end of the network. In this work, it is analyzed where to place the LSTM cells in these modern architectures to yield the best performance in the case of semantic video segmentation.


\section{RECURRENT NETWORK ARCHITECTURES FOR VIDEO SEGMENTATION}

	\begin{figure*}[tb]
		\centering
		\includegraphics[width=1.0\textwidth]{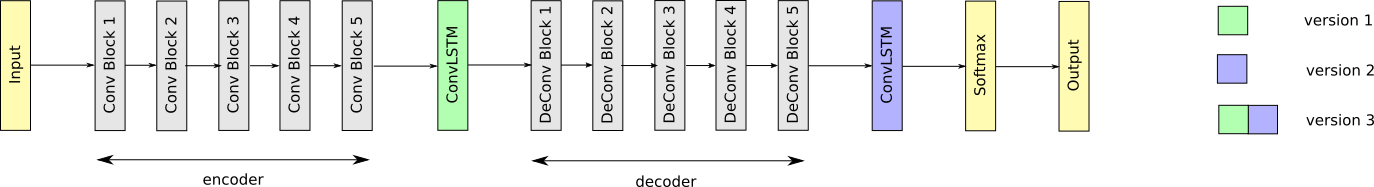}
		\caption{LSTM-SegNet network architecture: the gray boxes illustrate the original SegNet architecture, while the different positions of the ConvLSTM layers are illustrated in color. }
		\label{fig_LSTMSegNet_architecture}
	\end{figure*}
	
	\begin{figure*}[tb]
		\centering
		\includegraphics[width=1.0\textwidth]{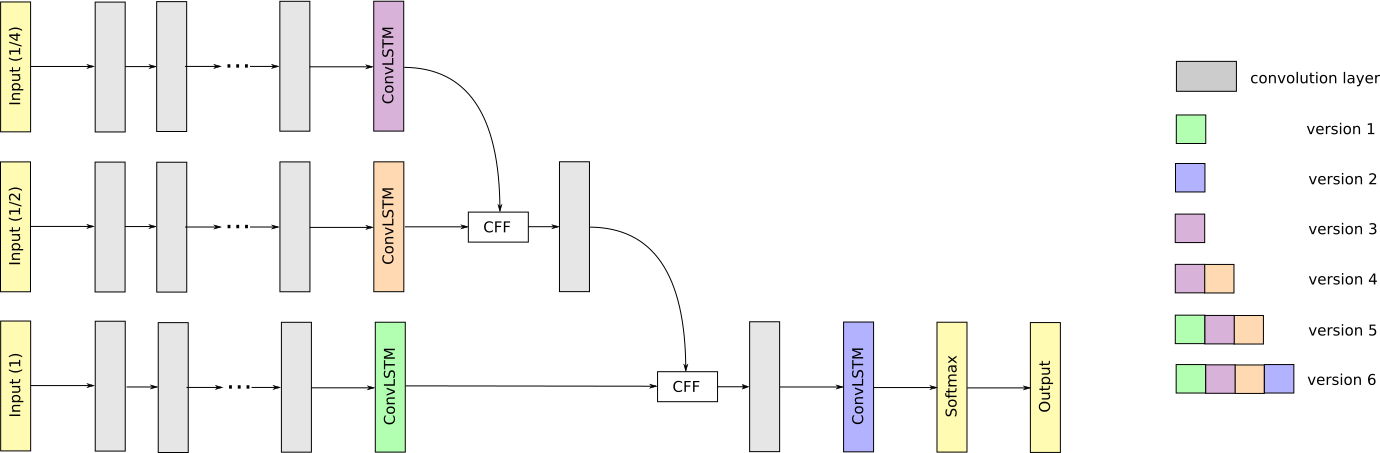}
		\caption{LSTM-ICNet network architecture: the gray boxes illustrate the original ICNet architecture, while the different positions of the ConvLSTM layers are illustrated in color. }
		\label{fig_LSTMICNetarchitecture}
	\end{figure*}

	In this section, different recurrent network architectures for the video segmentation task are presented. We take state-of-the-art segmentation approaches and extend them by convLSTM layers at different positions. 
	Concretely, SegNet and the ICNet are chosen as basic architectures, since they are real-time capable, which is an important issue for autonomous driving applications, and represent different network structures - the SegNet is an example of the classical encoder-decoder structure, and the ICNet is an example of a multi-branch approach.
	The extended SegNet is called hereinafter LSTM-SegNet and the extended ICNet LSTM-ICNet.
	In the following, the original network architectures and their extended versions of the considered approaches are described.

\subsection{LSTM-SegNet}

	The SegNet \cite{2015_Badrinarayanan_SegNet_ADeepConvolutionalEncoderDecoderARchitectureForImageSegmenatation} is one of the first real-time capable segmentation approaches, and is a typical representative of the classical encoder-decoder structure. 
	The encoder consists of convolutional and pooling layers of the VGG16 network \cite{Simonyan_2015_VeryDeepConvolutionalNetworksForLargeScaleImageRecognition}, and the decoder is a mirrored version of the encoder. Then, a softmax operation is applied to the output of the decoder to determine the class affiliation of each pixel. The advantage of the SegNet is that it is memory efficient, since it does not store the feature maps for the upsampling process in contrast to FCN \cite{Long_2015_FullyConvolutionalNetworksForSemanticSegmentation} but only its pooling indices. 
	
	The SegNet architecture is extended to the so called LSTM-SegNet by placing convLSTM layers at different positions in the network to find out which position is most suitable.  In previous works \cite{Nabavi_2018_FutureSemanticSegmentationWithConvolutionalLSTM, Valipour_2017_RecurrentFullyConvolutionalNetworksForVideoSegmentation, Yurdakul_2017_SemanticSegmentationOfRGBDVideosWithRecurrentFullyConvolutionalNeuralNetworks}, convLSTM layers are always located between the encoder and the decoder, hence, a convLSTM layer is also placed between the encoder and the decoder (version 1). The motivation for this position is that the encoder determines global image information. These information should not vary much between neighboring image frames so that storing and modifying these image features might avoid the flickering of features, and hence, might improve the segmentation result.
	Additionally, two different cases are considered. One possibility is to place a convLSTM layer at the end of the network directly in front of the softmax layer (version 2). This seems reasonable, because each frame can be processed independently without the influence of the previous ones so that the error can not be propagated through the sequence. At the end, the results of the current frame and the previous $N$ frames are combined by means of the recurrent structure. This corresponds to a temporal filtering of the segmentation result.	
	The last considered case (version 3) is a combination of the previous two versions. A convLSTM layer is placed between encoder and decoder, and another one is placed directly in front of the softmax layer.
	An overview of the different positions in the basic SegNet architecture is given in Fig. \ref{fig_LSTMSegNet_architecture}, where the gray boxes illustrate the (de)convolution blocks consisting of two to three convolutional layers, and the colored boxes represent different positions of the ConvLSTM layers in the network.

\subsection{LSTM-ICNet}

	Nowadays, the classical encoder-decoder architecture is not very popular any more for semantic segmentation tasks, since other network architectures containing several branches improve the segmentation performance. A typical representative of these new approaches is the ICNet \cite{Zhao_2017_ICNet_forRealTimeSemanticSegmentationOnHighResolutionImages}, which is very fast compared to other state-of-the-art approaches \cite{Chen_2017_DeepLab_SemanticImageSegmentationWithDeepConvolutionalNets_AtrousConvolution_andFullyConnectedCRFs, Zhao_2017_PyramidScenParsingNetwork}, and still delivers high accuracy. The speed-up is based on the usage of a highly downsampled version of the high-resolution input image. However, small image details get lost due to the downsampling. Hence, two further branches with greater resolution are used, which refine the prediction map of the lowest resolution. 
	Generally, the input image is downsampled twice, and input images of scale $1$, $1/2$ and $1/4$ are yielded. The image of the lowest resolution is fed into the top branch, which consists of a PSPNet based network architecture. The medium and high resolution images are processed in two further branches by several convolution layers. The feature map size of the higher resolutions is much smaller than the one of the lowest resolution to save computation time, since it is assumed that lower resolution layers contain the lost information.
	The output of the low and the medium resolution branches are fused by means of the Cascade Feature Fusion (CFF) layer \cite{Zhao_2017_PyramidScenParsingNetwork}. After a further convolution layer, the fused feature maps are concatenated with the output of the high resolution branch by means of a second CFF. The output of the CFF, which size is $1/4$ of the input image, is upsampled by means of interpolation and applying some convolution layers in two steps. Finally, a softmax operation is applied to predict the class of the image pixels.

	Analogously to the LSTM-SegNet, convLSTM layers are placed at different positions in order to find the best one. The ICNet does not consist of the classical encoder-decoder structure, and hence, convLSTM layers cannot be integrated between encoder and decoder. In contrast, convLSTM layers are placed at the end of each branch, before they are fused in the CFF layer (version 5). In version 2, a convLSTM layer is added before the softmax layer, which corresponds to a temporal filtering of the result similar to version 2 of the LSTM-SegNet. 
	Additionally, a combination of version 2 and version 5 is considered, where the convLSTM layers are placed at the end of each branch and in front of the softmax layer (version 6). Furthermore, it is analyzed if the performance increases by combining actual image information with temporal image information. For this, a convLSTM layer is placed only in one or two branches. In version 1, the convLSTM layers are added at the end of the high resolution branch to keep the local image information, and in version 3, the convLSTM layer is added at the end of the low resolution branch to keep global image information. Version 3 is further extended by additionally placing a convLSTM layer within the medium resolution branch (version 4). 
	Fig. \ref{fig_LSTMICNetarchitecture} shows an overview of the examined positions. The gray boxes illustrate the original ICNet structure, while the colored boxes correspond to possible positions of convLSTM layers. All in all, there are six possible positions where to add convLSTM layers in the ICNet. In the evaluation part, all of these possibilities are compared by means of state-of-the-art evaluation metrics.


\section{EVALUATION}

	In the previous sections, different possibilities were discussed where to best integrate convLSTM layers in existing networks. Now, these different possibilities are compared qualitatively and quantitatively in terms of pixelwise accuracy, mIoU, and inference time on the Cityscapes dataset \cite{cityscape_dataset}. The Cityscapes dataset consists of 5000 fine-annotated RGB images of size $1024 \times 2048$ recorded in 50 German cities. The dataset is split into 2975 images for training, 500 images for validation and 1525 images for testing. Since the ground-truth of the test images is not publicly available, the proposed approaches are evaluated on the validation set. The dataset contains 30 different classes, but similar to other works \cite{Zhao_2017_ICNet_forRealTimeSemanticSegmentationOnHighResolutionImages, Zhao_2017_PyramidScenParsingNetwork}, only 19 classes are used for training. For each image, the 19 previous and the 10 following images are also available. The time difference between two frames contains $60 ms$. We use sequences of four frames for training, i.e. the semantic segmentation map of frame $t$ is determined by means of the information of the last three frames $t-3$, $t-2$, and $t-1$.

	In this work, the focus is set on identical training conditions so that only the position of the convLSTM layers influences the result. Generally, the training parameters are chosen similar to \cite{Zhao_2017_ICNet_forRealTimeSemanticSegmentationOnHighResolutionImages}. The initial learning rate is set to $0.001$, 
	and decreased according to the poly learning rate policy. Furthermore, a momentum of $0.9$ and a weight decay of $0.0001$ is used, and the amount of iterations is set to $30k$.
	For the LSTM-SegNet and LSTM-ICNet, the same loss functions as in their original implementation are used. Note, that the loss is only determined for frame $t$, the errors of the previous frames $t-1, \dots,  t-3$ are ignored. To avoid overfitting, the training data are randomly mirrored and scaled with a ratio between $0.5$ and $2$. 
	The network parameters of the original approach and the extended versions are both initialized with the same pretrained networks. Moreover, all states of the added LSTM-cells are initialized with zero, i.e. the past is ignored completely. The kernel size of the convLSTM layers is set to $3 \times 3$, and the number of output channels is equal to the one of the previous layer. Each approach is implemented in Tensorflow \cite{tensorflow} and trained on a single Nvidia Titan X. Due to memory reasons, the batch size was set to one, knowing that the usage of greater batch sizes increases the performance further. For instance, the mIoU of the basic ICNet can be increased from $60.7\%$ (batch size 1) to $65.2\%$ (bath size 4) in our case. Note, that the origin ICNet approach proposed in \cite{Zhao_2017_ICNet_forRealTimeSemanticSegmentationOnHighResolutionImages} achieves a mIoU of $67.7\%$ (bath size 16). This difference might be a consequence of different batch sizes and due to implementation reasons.  
	The results of LSTM-SegNet and LSTM-ICNet are discussed in the following sections.

\subsection{LSTM-SegNet}

\begin{figure*}[p]
		\centering
		\includegraphics[width=0.9\textwidth]{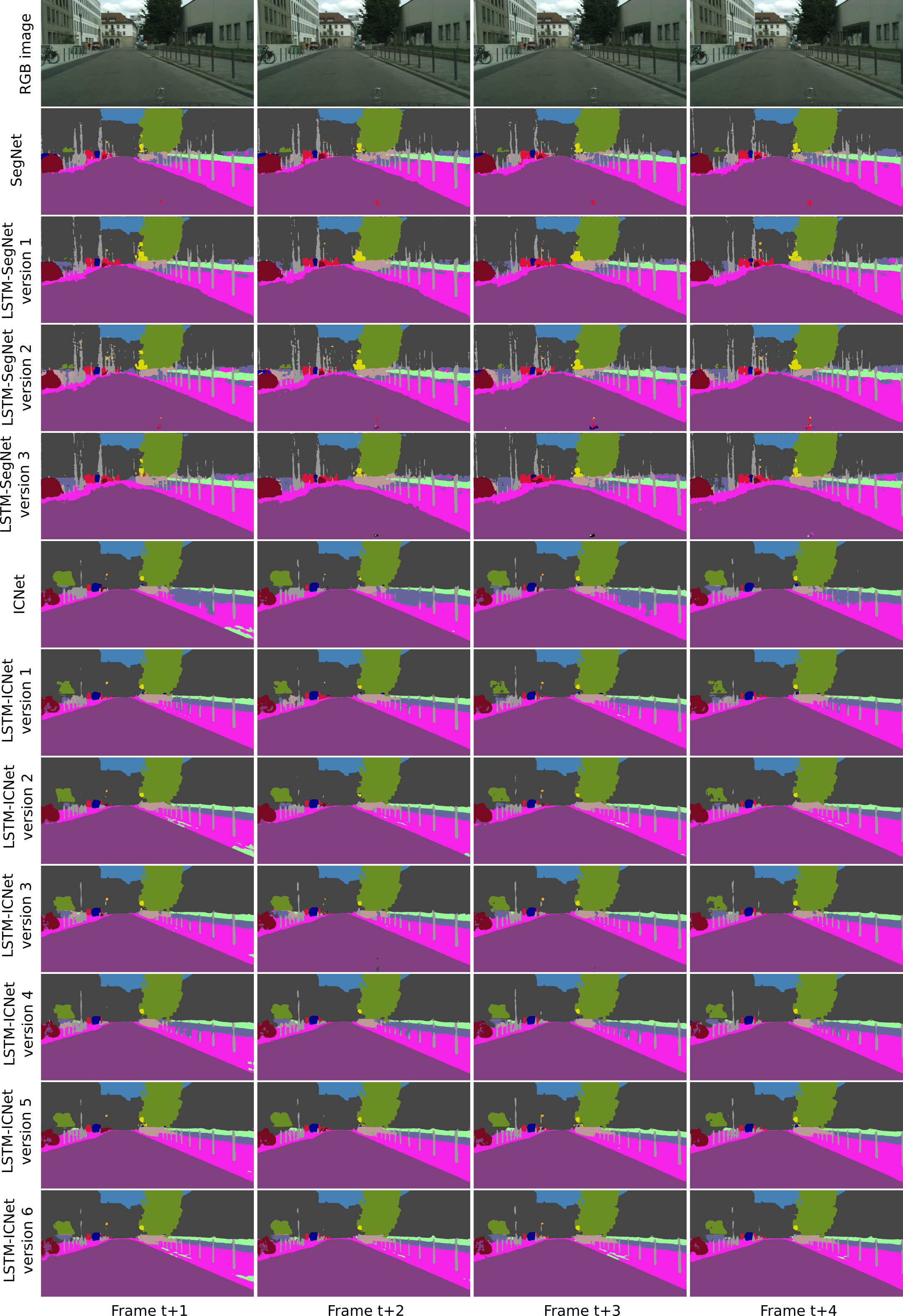}
		\caption{Qualitative results of the proposed approaches over a video clip of the Cityscapes dataset. The first row contains the RGB images, the next four rows the segmentation map of the SegNet-based approaches and the last seven rows the results of the ICNet-based approaches.}
		\label{fig_LSTMICNetexamples1}
	\end{figure*}

	The three described LSTM-SegNet approaches of the previous section are now evaluated, and compared with the original SegNet approach. The resolution of the input images is reduced to $256 \times 512$ due to memory and time reasons. The results of the considered approaches can be found in Table \ref{table_example_Cityscapes}. It turns out that the LSTM-based versions outperform the original SegNet by up to $1.3\%$ in terms of mIoU, while all approaches perform similarly in terms of pixelwise accuracy. LSTM-SegNet version 2 yields the best results compared to all LSTM-based approaches, and is by about $0.7\%$ better than version 1 in terms of mIoU, which corresponds to the approaches of previous works \cite{Nabavi_2018_FutureSemanticSegmentationWithConvolutionalLSTM, Valipour_2017_RecurrentFullyConvolutionalNetworksForVideoSegmentation, Yurdakul_2017_SemanticSegmentationOfRGBDVideosWithRecurrentFullyConvolutionalNeuralNetworks}.   	
	Version 3, which is a combination of version 1 and version 2, performs worst of the LSTM-based versions. It seems that training multiple convLSTM layers is more challenging than training only one.   
	The LSTM-based approaches also perform visually better, e.g. in Fig. \ref{fig_LSTMICNetexamples1}, the LSTM-based approaches partly detect the sidewalk on the left side and the wall on the right in contrast to the original SegNet approach.
	Furthermore, the proposed approaches are also evaluated based on the inference time. According to the results listed in Table \ref{table_example_Cityscapes}, the original SegNet architecture is at least twice as fast as the LSTM-based approaches. The reason is that convLSTM layers are more time intensive than convolutional layers, since several convolutions have to be applied in the convLSTM layer for determining its internal states. LSTM-SegNet version 2 takes longer than version 1, because the input size of the processed feature map is much larger. Version 3 is a combination of version 1 and version 2, and hence, its inference time is the greatest one.

\subsection{LSTM-ICNet}

	\begin{table}[t]
		\caption{Evaluation on Cityscapes}
		\centering
		\begin{center}
			\begin{tabular}{|c||c|c|c|}
				\hline
				approach & accuracy(\%) & mIoU(\%) & inference time \Tstrut \Bstrut \\
				\hline \hline \Tstrut
				SegNet & $84.53$ & $38.21$ & $ 34.80 ms$ \Mstrut\\
				ICNet & $92.66$ & $60.63$ & $ 60.91 ms$ \Bstrut \\
				\hline \Tstrut
				LSTM-SegNet version 1  & $84.21$ & $38.79$ & $ 74.44 ms$ \Mstrut \\
				LSTM-SegNet version 2  & $84.41$ & $39.54$ & $ 79.35 ms$ \Mstrut \\
				LSTM-SegNet version 3  & $83.03$ & $38.70$ & $ 81.42 ms$ \Bstrut \\
				\hline \Tstrut
				LSTM-ICNet version 1  & $93.03$ & $61.55$ & $ 74.46 ms$ \Mstrut \\
				LSTM-ICNet version 2  & $92.99$ & $61.96$ & $74.80 ms$ \Mstrut \\
				LSTM-ICNet version 3  & $92.92$ & $61.47$ & $69.52 ms$ \Mstrut \\
				LSTM-ICNet version 4  & $92.81$ & $60.95$ & $71.59 ms$ \Mstrut \\
				LSTM-ICNet version 5  & $92.92$ & $\mathbf{62.28}$ & $79.17 ms$ \Mstrut \\
				LSTM-ICNet version 6  & $\mathbf{93.04}$ & $61.52$ & $83.16 ms$ \Bstrut \\
				\hline
			\end{tabular}
		\end{center}
		\label{table_example_Cityscapes}
		\vspace{-5mm}
	\end{table}

	In this section, the LSTM-ICNet is evaluated on the Cityscapes dataset. Similar to the original implementation, the full resolution images of size $1024 \times 2048$ are used for training and evaluation. 
	Visually, the LSTM-based approaches reduce the amount of flickering objects, and detect image areas more coherently. For example, in Fig \ref{fig_LSTMICNetexamples1}, the sidewalk is recognized more continuously.
	Moreover, fine structures, such as poles, are detected better, although they are not always classified correctly. The results of the LSTM-based approaches do not distinguish much. Nevertheless, small differences can be recognized. For instance, \mbox{version 1} recognizes fine structures better, and the edges are usually smoother. In contrast, version 3 has more problems with fine structures. However, it classifies large image areas better, since the network stores global information. \mbox{Version 2} corresponds to a temporal filtering of the segmentation map. If the error occurs only in a few frames, the error can be often remedied, otherwise, the error still appears. The other considered versions are combinations of the first three versions, and share their advantages and disadvantages, as the results show.
	
	The proposed approaches are also evaluated quantitatively in terms of pixelwise accuracy and mIoU. The results can be found in Table \ref{table_example_Cityscapes}. The results show that the LSTM-based approaches outperform the ICNet by at least $0.3\%$ in terms of pixelwise accuracy and from about $1\%$ up to $1.6\%$ in terms of mIoU, which shows that adding temporal information to the system does improve the performance of the segmentation approach. 
	The LSTM-based approaches have similar accuracies, while version 6 delivers the highest accuracy. A more meaningful evaluation metric is mean IoU (mIoU), since the class frequency is also considered. 	
	The best mIoU value is yielded by version 5. 
	Version 2, which corresponds to a temporal filtering of the results, also performs well analogously to the SegNet-based approaches, and achieves the best results of all approaches using only one convLSTM layer.
	Combining the different LSTM positions should expectedly improve the performance. However, this does not always hold true according to the results, since the advantages and disadvantages of the different positions might cancel each other out. Furthermore, training several LSTM layers is more challenging from experience, and hence, the optimization problem convergence worsens.

	Similar to the SegNet-based architectures, the inference time of the pure CNN-based ICNet is lower than of the LSTM-based approaches,
	since applying a convLSTM layer is more time-intensive than applying a convolutional layer. The fastest LSTM-ICNet version is version 3 and only takes about $10ms$ longer.  
	The reason is that the convLSTM layers have got the same kernel and channel size, but the size of their feature maps varies. Hence, applying the convLSTM layer of the medium and the low resolution branch takes only a quarter and a sixteenth of the computation time of the high resolution branch, respectively. The computation time of version 1 and version 2 is similar since their convLSTM layers have identical dimensions. The slowest inference time shows in version 6, because it is a combination of the other versions.
	Each of the proposed approaches takes less than $100ms$, which corresponds to a frame rate of more than 10 frames per second, so that the LSTM-based approaches can be still used in real-time capable applications.
	In this work, we only use a kernel size of $3 \times 3$ for the convLSTM layers. A greater kernel size increases the spatiotemporal area, and improves the performance of the proposed algorithms further, as experiments show. However, the inference time increases enormously, so that the real-time capability condition cannot be fulfilled. Hence, this case is not considered in this paper.

	\begin{table}[t]
		\caption{Evaluation on Virtual Kitti}
		\centering
		\begin{center}
			\begin{tabular}{|c||c|c|}
				\hline
				approach & accuracy(\%) & mIoU(\%)  \Tstrut \Bstrut \\
				\hline \hline \Tstrut
				Conv-RGB$^{\mathrm{*}}$ \cite{Yurdakul_2017_SemanticSegmentationOfRGBDVideosWithRecurrentFullyConvolutionalNeuralNetworks} & $80.01$ & -- \Mstrut \\
				LSTM-RGB$^{\mathrm{*}}$ \cite{Yurdakul_2017_SemanticSegmentationOfRGBDVideosWithRecurrentFullyConvolutionalNeuralNetworks} & $79.95$ & -- \Mstrut \\
				ICNet & $85.32$ & $42.77$  \Mstrut \\
				LSTM-ICNet version 1  &  $86.36$ & $44.48$ \Mstrut \\
				LSTM-ICNet version 2  &  $86.32$ & $44.60$ \Mstrut \\
				LSTM-ICNet version 3  &  $86.15$ & $44.40$ \Mstrut \\
				LSTM-ICNet version 4  &  $86.72$ & $44.84$ \Mstrut \\
				LSTM-ICNet version 5  &  $86.40$ & $44.58$ \Mstrut \\
				LSTM-ICNet version 6  &  $\mathbf{86.96}$ & $\mathbf{45.59}$ \Bstrut \\
				\hline \Tstrut
				ICNet$^{\mathrm{**}}$  & $92.60$ & $58.44$ \Mstrut \\ 
				LSTM-ICNet version 1$^{\mathrm{**}}$  & $92.74$ & $59.04$ \Mstrut \\
				LSTM-ICNet version 2$^{\mathrm{**}}$  & $93.01$ & $59.71$ \Mstrut \\
				LSTM-ICNet version 3$^{\mathrm{**}}$  & $92.58$ & $58.94$ \Mstrut \\
				LSTM-ICNet version 4$^{\mathrm{**}}$  & $92.77$ & $59.57$ \Mstrut \\
				LSTM-ICNet version 5$^{\mathrm{**}}$  & $92.96$ & $60.19$ \Mstrut \\
				LSTM-ICNet version 6$^{\mathrm{**}}$  & $\mathbf{93.07}$ & $\mathbf{60.50}$ \Bstrut \\
				\hline
				\multicolumn{3}{l}{$^{\mathrm{*}}$ results according to original paper (\cite{Yurdakul_2017_SemanticSegmentationOfRGBDVideosWithRecurrentFullyConvolutionalNeuralNetworks})} \Tstrut \\
				\multicolumn{3}{l}{$^{\mathrm{**}}$ trained on images with full resolution ($375 \times 1242$)} \Tstrut \\
			\end{tabular}
		\end{center}
		\vspace{-5mm}
		\label{table_exampleVirtuellKitti}
	\end{table}

	Finally, the  origin ICNet and the proposed LSTM-ICNet versions are compared with other state-of-the-art methods.
	There only exist a few current video segmentation approaches as described in Section \ref{section_related_work}, hence, the considered approaches are only compared with the work of Yurdakul et al. \cite{Yurdakul_2017_SemanticSegmentationOfRGBDVideosWithRecurrentFullyConvolutionalNeuralNetworks} on the virtual Kitti dataset \cite{virtualKittiDataset}. The virtual Kitti dataset consists of 50 photo-realistic synthetic video sequences and the corresponding semantic labels for each frame, which are 21260 images in total.
	Analogously to \cite{Yurdakul_2017_SemanticSegmentationOfRGBDVideosWithRecurrentFullyConvolutionalNeuralNetworks}, the resolution of the input images is reduced to $224 \times 224$ without preserving their aspect ratios. The training set consists of the first halves of all variations and the test set of the second halves, which corresponds to setup 2 in \cite{Yurdakul_2017_SemanticSegmentationOfRGBDVideosWithRecurrentFullyConvolutionalNeuralNetworks}. 
	The training conditions are identical to the ones of the previous sections, only that a batch size of two is used. 
	The results (see Table \ref{table_exampleVirtuellKitti}) show that the proposed approaches outperform the LSTM-RGB net \cite{Yurdakul_2017_SemanticSegmentationOfRGBDVideosWithRecurrentFullyConvolutionalNeuralNetworks} by at least $5\%$ in terms of accuracy. 
	Additionally, the LSTM-RGB net performs similarly to its pure CNN-based variant (Conv-RGB). In contrast, our LSTM-based approaches achieve better results by at least $1\%$ percent compared to the origin ICNet. 
	For the sake of completeness, the  origin ICNet and the different LSTM-ICNet are also trained and evaluated with the original image size ($375 \times 1242$) using a batch size of one again. The results are also shown in \mbox{Table \ref{table_exampleVirtuellKitti}}.


\section{CONCLUSION}

	In this paper, state-of-the-art semantic segmentation approaches have been extended by convLSTMs to also consider image information of previous frames. Different positions of the LSTM cells were investigated with different network architectures such as the encoder-decoder structure and the multi-branch systems and evaluated on the Cityscapes dataset. It turned out that the convLSTM-based approaches outperform the original approaches by up to $1.6\%$, while its real-time capability can be still guaranteed, i.e. the inference time of the LSTM-based approaches is less than $100ms$. The experiments show that the best position of the convLSTM layers is directly in front of the softmax layer in case of the encoder-decoder architecture, and at the end of the high-resolution branch in case of the multi-branch architecture. We further found out that combining different positions does not necessarily improve the performance.


\bibliography{/home/andreas/Documents/Literatur/Jabref-Datebase/Literatur_Promotion}
\bibliographystyle{plain}

\end{document}